\documentclass[tbtags]{amsart2000}
\usepackage{amssymb}
\usepackage{calc}

\calclayout
\let\Large\large

\usepackage{linguex}
\renewcommand{\refdash}{}
\makeatletter
\let\c@ExNo\c@equation
\let\theExNo\theequation
\renewcommand{\theSubExNo}{\hbox{\if@noftnote\arabic{ExNo}\else
   \roman{FnExNo}\fi\refdash\alph{SubExNo}}}
\renewcommand{\theSubSubExNo}{%
   \hbox{\if@noftnote\arabic{ExNo}\else\roman{FnExNo}\fi
          \refdash\alph{SubExNo}\refdash\if@noftnote\roman{SubSubExNo}%
                          \else\arabic{SubSubExNo}\fi}}
\makeatother

\newcommand{\CCSHANdenotation}[1]{\text{\scshape\lowercase{#1}}}
\newcommand{\CCSHANA          }{\CCSHANdenotation{a}}
\newcommand{\CCSHANBeat       }{\CCSHANdenotation{beat}}
\newcommand{\CCSHANDonkey     }{\CCSHANdenotation{donkey}}
\newcommand{\CCSHANEvery      }{\CCSHANdenotation{every}}
\newcommand{\CCSHANFarmer     }{\CCSHANdenotation{farmer}}
\newcommand{\CCSHANHe         }{\CCSHANdenotation{he}}
\newcommand{\CCSHANIt         }{\CCSHANdenotation{it}}
\newcommand{\CCSHANLove       }{\CCSHANdenotation{love}}
\newcommand{\CCSHANMan        }{\CCSHANdenotation{man}}
\newcommand{\CCSHANOwn        }{\CCSHANdenotation{own}}
\newcommand{\CCSHANShe        }{\CCSHANdenotation{she}}
\newcommand{\CCSHANWhistle    }{\CCSHANdenotation{whistle}}
\newcommand{\CCSHANWho        }{\CCSHANdenotation{who}}
\newcommand{\CCSHANWitp       }{\CCSHANdenotation{witp}}

\newcommand{\CCSHANeqish}[1]{\mathrel{\makebox[0pt]{\hphantom{$=$}\hss$#1$\hss}\hphantom{=}}}
\newcommand{\CCSHANeqishcolon}{\CCSHANeqish{:}}

\newcommand{\CCSHANphrase}[1]{\emph{#1}}
\newcommand{\CCSHANfun}[1]{\mathopen{\lambda\mathord{#1}.\,}}
\newcommand{\CCSHANtoF}{\mathbin\rightarrow}
\newcommand{\CCSHANIn }{\mathbin\vartriangleright}
\newcommand{\CCSHANOut}{\mathbin\ltimes}
\newcommand{\CCSHANgIn }{g^{\CCSHANIn }}
\newcommand{\CCSHANgOut}{g^{\CCSHANOut}}
\newcommand{\CCSHANstuff}{\text{--}}
\newcommand{\CCSHANcomma}{\text{, }}
\newcommand{\CCSHANsref}[1]{\S\ref{#1}}
\newcommand{\CCSHANForall}[1]{\mathopen{\forall\mathord{#1}.\,}}
\newcommand{\CCSHANExists}[1]{\mathopen{\exists\mathord{#1}.\,}}
\newcommand{\CCSHANImplies}{\mathrel\Rightarrow}
\newcommand{\CCSHANIndex}[1]{\ensuremath{_{\text{#1}}}}
\DeclareMathOperator{\CCSHANSet}{Set}

\begin{document}
\thispagestyle{empty}
\centerline{\normalfont\Large\bfseries A Variable-Free Dynamic Semantics}
\vspace{2.3ex plus.2ex}
\centerline{\normalfont Chung-chieh Shan}
\centerline{\normalfont Harvard University}
\vspace{3.5ex plus1ex minus.2ex}
{\let\thefootnote\relax\footnotetext{Thanks to Pauline Jacobson and Stuart
Shieber for discussion and comments.  This work is supported by the
United States National Science Foundation under Grant IRI-9712068.}}

I propose a variable-free treatment of dynamic semantics.  By ``dynamic
semantics'' I mean analyses of donkey sentences (\CCSHANphrase{Every farmer who
owns a donkey beats it}) and other binding and anaphora phenomena in
natural language where meanings of constituents are updates to information
states, for instance as proposed by Groenendijk and
Stokhof~\cite{groenendijk-dynamic}.  By ``variable-free'' I mean
denotational semantics in which functional combinators replace variable
indices and assignment functions, for instance as advocated by
Jacobson~\cite{jacobson-variable-free,jacobson-paycheck}.

The new theory presented here achieves a compositional treatment of dynamic
anaphora that does not involve assignment functions, and separates the
combinatorics of variable-free semantics from the particular linguistic
phenomena it treats.  Integrating variable-free semantics and dynamic
semantics gives rise to interactions that make new empirical predictions,
for example ``donkey weak crossover''~effects.

\section{Decomposing dynamism}

Dynamic semantics combines \emph{nondeterminism}, \emph{input}, and
\emph{output} to interpret discourse fragments such as~\eqref{e:a+he} in a
process informally described in~\eqref{e:imperative}.
\ex.\label{e:a+he}
    A man walks in the park. He whistles.\par
\ex.\label{e:imperative}
    \emph{Nondeterministically} select a man $x$.\\
    \emph{Output} $x$ as a candidate antecedent for future anaphora.\\
    Check to make sure that $x$ walks in the park; if not, abort execution.\\
    \emph{Input} a previously encountered candidate antecedent $y$.\\
    Check to make sure that $y$ whistles; if not, abort execution.\par
In this section, I analyze each aspect in turn.  In subsequent sections, I
will then review the empirical and theoretical advantages gained in my
variable-free treatment.

\subsection{Nondeterminism}
\label{s:nondeterminism}

In the first half of~\eqref{e:a+he}, \CCSHANphrase{a man} nondeterministically
selects a man, who is then tested for the property \CCSHANphrase{walks in the
park}.  If any choice of a man passes the test, the sentence is true;
otherwise, it is false.  Denotationally, I model nondeterminism by letting
phrases denote sets of what they traditionally denote in Montague grammar.
For example, I let noun phrases denote not individuals but sets of
individuals: \CCSHANphrase{John} denotes the singleton set containing John, and
\CCSHANphrase{a man} the set of all men.  Formally, I assign the type $\CCSHANSet(e)$
(or $e\CCSHANtoF t$), rather than $e$, to noun phrases such as \CCSHANphrase{John} and
\CCSHANphrase{a man}.  Here $e$ is the base type of individuals and $t$ is the
base type of truth values.

Now write $1$ for the \emph{unit type}, the identity for the binary type
constructor~$\times$ for product types.  This type can be thought of as a
singleton set, say $\{*\}$.  Note that $\{*\}$ has two subsets, namely
$\{*\}$ and $\{\}$.  I treat these two subsets as true and false,
respectively, thus establishing an isomorphism between the types $t$ and
$\CCSHANSet(1)$.  I then assign to \CCSHANphrase{walks in the park} the type
$\CCSHANSet(e\CCSHANtoF\CCSHANSet(1))$ rather than $e\CCSHANtoF t$.  We can think of a property as
a nondeterministic function that maps each $e$ to a nondeterministic $1$,
returning either $*$ or nothing---that is, either $\{*\}$ or $\{\}$.

In general, nondeterminism can be added to any Montague grammar by
replacing each semantic type $\tau$ with a transformed type
$\CCSHANSet(\lfloor\tau\rfloor)$, where $\lfloor\CCSHANstuff\rfloor$ is a map from
types to types recursively defined by
\begin{subequations}
\begin{align}
    \lfloor\tau\rfloor
        &= \tau
        &&\text{for any base type $\tau$,}\\
    \lfloor\tau_1\CCSHANtoF\tau_2\rfloor
        &= \lfloor\tau_1\rfloor\CCSHANtoF\CCSHANSet(\lfloor\tau_2\rfloor)
        &&\text{for any types $\tau_1$ and $\tau_2$.}
\end{align}
\end{subequations}
My type for properties $\CCSHANSet(e\CCSHANtoF\CCSHANSet(1))$, for example, is precisely
$\CCSHANSet(\lfloor e\CCSHANtoF 1\rfloor)$.  Once every type is transformed, it is
straightforward to specify how semantic values compose, either by adding a
composition method (as in Hamblin's interrogative
semantics~\cite{hamblin-questions}) or by adding a type-shift operation.
In programming language terms, I am using a lambda calculus that is
\emph{impure} because it incorporates call-by-value nondeterminism, as has
been detailed by others~\cite[\S8]{wadler-comprehending}.

For brevity, I will henceforth write untransformed types in place of
transformed ones.  For example, I will write $e$ rather than $\CCSHANSet(e)$ for
the type of an individual, $e\CCSHANtoF 1$ rather than $\CCSHANSet(e\CCSHANtoF\CCSHANSet(1))$ for
the type of a property, and $e\CCSHANtoF e\CCSHANtoF 1$ rather than
$\CCSHANSet(e\CCSHANtoF\CCSHANSet(e\CCSHANtoF\CCSHANSet(1)))$ for the type of a two-place relation.
Intuitively, a type $\tau_1\CCSHANtoF\tau_2$ is henceforth to be interpreted as a
relation between $\tau_1$ and $\tau_2$, or equivalently, a function from
$\tau_1$ to the power set of $\tau_2$.  Accordingly, if $f$ is of type
$\tau_1\CCSHANtoF\tau_2$ and $x$ is of type $\tau_1$, then the term $f(x)$, of
type $\tau_2$, is to be interpreted as the image of the set $x$ under the
relation $f$.

We can now derive \CCSHANphrase{A man walks in the park}:
\begin{gather*}
    \CCSHANA : (e\CCSHANtoF 1)\CCSHANtoF e = \CCSHANfun{p} \bigl\{\, v \mid *\in p(v) \,\bigr\}, \\
    \CCSHANMan : e\CCSHANtoF 1, \qquad
    \CCSHANWitp : e\CCSHANtoF 1, \qquad
    \CCSHANWitp(\CCSHANA(\CCSHANMan)) : 1.
\end{gather*}

\subsection{Input}

In the second half of~\eqref{e:a+he}, a man \CCSHANphrase{he} is determined by
the discourse context, who is then tested for the property
\CCSHANphrase{whistles}.  The central idea of variable-free semantics is to model
dependence on discourse context by letting phrases denote functions from
\emph{inputs} to what they traditionally denote in Montague grammar.  For
example, \CCSHANphrase{he} will denote the identity function over men, and
\CCSHANphrase{he whistles} the function mapping each man to whether he whistles.

To restate this idea formally in terms of types, I introduce a new binary
type constructor $\CCSHANIn$ (``in'').  The type $\sigma\CCSHANIn\tau$ is like
$\sigma\CCSHANtoF\tau$ in that they may have the same models, namely functions
from $\sigma$ to $\tau$.  I use for both kinds of types the same
$\CCSHANfun{\CCSHANstuff}\CCSHANstuff$ notation for abstraction and $\CCSHANstuff(\CCSHANstuff)$ notation
for application, but distinguish between them so that, for example, a value
of type $(a\CCSHANtoF b)\CCSHANtoF c$ cannot apply directly to one of type $a\CCSHANIn b$.
(This is equivalent to how, in Jacobson's formulation, syntactic categories
regulate semantic combination to stop \CCSHANphrase{loves him} from applying to
\CCSHANphrase{Mary} \cite[\S2.2.1.1]{jacobson-variable-free}.)  By convention,
all binary type constructors associate to the right.

I assume that \CCSHANphrase{whistles} denotes some property $\CCSHANWhistle : e \CCSHANtoF
1$, and let \CCSHANphrase{he} denote
\begin{equation*}
\CCSHANHe : e \CCSHANIn e = \CCSHANfun{v} v.
\end{equation*}
Because $\CCSHANHe$ does not have type $e$, the property $\CCSHANWhistle$ cannot apply
to it directly.  I follow Jacobson in introducing a type-shift operation
\begin{equation}
\CCSHANgIn : (\alpha\CCSHANtoF\beta)\CCSHANtoF(\sigma\CCSHANIn\alpha)\CCSHANtoF(\sigma\CCSHANIn\beta)
     = \CCSHANfun{f} \CCSHANfun{v} \CCSHANfun{s} f(v(s)).
\end{equation}
We can now derive \CCSHANphrase{he whistles}:
\begin{equation*}
\CCSHANgIn(\CCSHANWhistle)(\CCSHANHe) : e\CCSHANIn 1 = \CCSHANfun{v} \CCSHANWhistle(v).
\end{equation*}

For phrases containing more than one pronoun, for example \CCSHANphrase{he loves
her}, I generalize the type-shift operation $\CCSHANgIn$ to a family of
operations $\CCSHANgIn_{i,j} = G^i(I^j(\CCSHANgIn))$ for non-negative integers $i$
and~$j$, that is,
\[
    \CCSHANgIn_{0,0}   = \CCSHANgIn,                 \qquad
    \CCSHANgIn_{0,j+1} = I(\CCSHANgIn_{0,j}),        \qquad
    \CCSHANgIn_{i+1,j} = G(\CCSHANgIn_{i,j}),
\]
where the ``composition'' function $G$ and the ``insertion'' function $I$
are defined by
\begin{align}
G &\CCSHANeqishcolon (\alpha\CCSHANtoF\alpha')\CCSHANtoF(\tau\CCSHANtoF\alpha)\CCSHANtoF(\tau\CCSHANtoF\alpha')
   = \CCSHANfun{g}\CCSHANfun{f}\CCSHANfun{v} g(f(v)),\displaybreak[0]\\
I &\CCSHANeqishcolon \bigl((\alpha\CCSHANtoF\beta)\CCSHANtoF(\alpha'\CCSHANtoF\beta')\bigr)\CCSHANtoF
     \bigl((\alpha\CCSHANtoF\tau\CCSHANtoF\beta)\CCSHANtoF(\alpha'\CCSHANtoF\tau\CCSHANtoF\beta')\bigr)\\
  &= \CCSHANfun{g}\CCSHANfun{f}\CCSHANfun{v'}\CCSHANfun{x} g (\CCSHANfun{v} f(v)(x))(v').\notag
\end{align}
Assuming that \CCSHANphrase{she} denotes $\CCSHANShe = \CCSHANHe$ and \CCSHANphrase{loves} denotes
$\CCSHANLove : e\CCSHANtoF e\CCSHANtoF 1$, we can now derive two denotations for \CCSHANphrase{he
loves her} with opposite scoping:
\begin{align*}
    \CCSHANgIn_{1,0}\bigl(\CCSHANgIn_{0,1}(\CCSHANLove)\bigr)(\CCSHANShe)(\CCSHANHe) &: e\CCSHANIn e\CCSHANIn 1 = \CCSHANfun{u} \CCSHANfun{v} \CCSHANLove(v)(u),\\
    \CCSHANgIn_{0,1}\bigl(\CCSHANgIn_{1,0}(\CCSHANLove)\bigr)(\CCSHANShe)(\CCSHANHe) &: e\CCSHANIn e\CCSHANIn 1 = \CCSHANfun{v} \CCSHANfun{u} \CCSHANLove(v)(u).
\end{align*}
My generalization here of $\CCSHANgIn$ to handle multiple pronouns differs from
Jacobson's, which does not posit $\CCSHANgIn_{i,j}$ for $i > 0$ and only
generates the second scoping.  The first scoping will be crucial as we
consider output and binding below---typically, we need to use $\CCSHANgIn_{1,0}$
before the binding operation $z$, defined in~\eqref{e:z}, can apply.

\subsection{Output}

Informally speaking, \CCSHANphrase{a man} can bind \CCSHANphrase{he} in~\eqref{e:a+he}
by introducing a new discourse referent, i.e., a new candidate antecedent
for future anaphora.  I model this kind of addition to discourse context by
letting phrases denote cartesian products between \emph{outputs} and what
they traditionally denote in Montague grammar.  For example, \CCSHANphrase{a man}
will denote the set of all pairs $\langle v,v\rangle$ where $v$ is a man,
and \CCSHANphrase{a man walks in the park} the set of all pairs $\langle
v,*\rangle$ where $v$ is a man who walks in the park.

Formally, I introduce a new binary type constructor $\CCSHANOut$ (``out'').  The
type $\sigma\CCSHANOut\tau$ is like $\sigma\times\tau$ in that they may have
the same models, namely pairs between $\sigma$ and~$\tau$.  I will use
for both kinds of types the same $\langle\CCSHANstuff,\CCSHANstuff\rangle$ notation for
pairs, but distinguish between them so that, for example, a value of type
$(a\times b)\CCSHANtoF c$ cannot apply to another of type $a\CCSHANOut b$.
For simplicity, I treat as equivalent the isomorphic types
\begin{equation*}
    1\CCSHANOut\tau \quad\text{and}\quad \tau
\end{equation*}
for any type $\tau$, and the isomorphic types
\begin{equation*}
    (\sigma_1\times\sigma_2)\CCSHANOut\tau \quad\text{and}\quad \sigma_1\CCSHANOut(\sigma_2\CCSHANOut\tau)
\end{equation*}
for any types $\tau$, $\sigma_1$, and $\sigma_2$.

As with input, I introduce a type-shift operation
\begin{equation}
\CCSHANgOut : (\alpha\CCSHANtoF\beta)\CCSHANtoF(\sigma\CCSHANOut\alpha)\CCSHANtoF(\sigma\CCSHANOut\beta)
      = \CCSHANfun{f} \CCSHANfun{\langle s,v\rangle} \langle s, f(v)\rangle.
\end{equation}
I then generalize $\CCSHANgOut$ to a family of type-shift operations
$\CCSHANgOut_{i,j} = G^i(I^j(\CCSHANgOut))$ for non-negative integers $i$ and~$j$.

Recall from~\CCSHANsref{s:nondeterminism} that true and false are just nonempty
and empty sets, respectively.  Under this view, it is easy to define a
concatenation function that conjoins the truth conditions of two discourse
fragments:
\begin{equation*}
\mathord{;} : 1\CCSHANtoF 1\CCSHANtoF 1 = \CCSHANfun{*} \CCSHANfun{*} *.
\end{equation*}
(Syntactically, I assume that the first argument to $;$ is the second of
the two fragments to be conjoined, and vice versa.) Revising the denotation
we specified earlier for \CCSHANphrase{a}, we can now derive the reading
of~\eqref{e:a+he} where \CCSHANphrase{a man} does not bind \CCSHANphrase{he}:
\begin{gather*}
    \CCSHANA : (e\CCSHANOut e \CCSHANtoF \sigma\CCSHANOut 1)\CCSHANtoF \sigma\CCSHANOut e
       = \CCSHANfun{p} \bigl\{\, \langle s,v\rangle
                      \mid \langle s,*\rangle \in p(\langle v,v\rangle) \,\bigr\},\\
    \CCSHANgOut_{0,0}(\CCSHANWitp)(\CCSHANA(\CCSHANgOut_{0,0}(\CCSHANMan))) : e\CCSHANOut 1,\\
    \CCSHANgOut_{1,0}(\CCSHANgIn_{0,1}(\mathord{;}))
        \bigl(\CCSHANgIn_{0,0}(\CCSHANWhistle)(\CCSHANHe)\bigr)
        \bigl(\CCSHANgOut_{0,0}(\CCSHANWitp)(\CCSHANA(\CCSHANgOut_{0,0}(\CCSHANMan)))\bigr)
        : e\CCSHANOut e\CCSHANIn 1.
\end{gather*}

For binding to take place, we need to feed outputs produced by semantically
higher arguments into inputs solicited by semantically lower arguments.  To
implement this, I define one last type-shift operation
\begin{equation}
\label{e:z}
\begin{split}
z &\CCSHANeqishcolon
     \bigl(          \alpha  \CCSHANtoF (\sigma\CCSHANOut\beta) \CCSHANtoF \gamma\bigr) \CCSHANtoF
     \bigl((\sigma\CCSHANIn\alpha) \CCSHANtoF (\sigma\CCSHANOut\beta) \CCSHANtoF \gamma\bigr)\\
  &= \CCSHANfun{f} \CCSHANfun{v} \CCSHANfun{\langle s,u\rangle} f(v(s))(\langle s,u\rangle)
\end{split}
\end{equation}
and derive from it a family of type-shift operations $z_{i,j} =
G^i(I^j(z))$ for non-negative integers $i$ and~$j$.  We can now derive the
reading of~\eqref{e:a+he} where \CCSHANphrase{a man} does bind~\CCSHANphrase{he}:
\[
    z(\CCSHANgIn_{1,0}(\mathord{;}))
        \bigl(\CCSHANgIn_{0,0}(\CCSHANWhistle)(\CCSHANHe)\bigr)
        \bigl(\CCSHANgOut_{0,0}(\CCSHANWitp)(\CCSHANA(\CCSHANgOut_{0,0}(\CCSHANMan)))\bigr) : e\CCSHANOut 1.
\]

\section{Empirical payoffs}

Many variable-free analyses of empirical facts carry over in spirit to the
variable-free dynamic semantics presented here, with extended coverage over
dynamic phenomena.  In this section, I give some simple examples that
center around the classical donkey sentence~\eqref{e:donkey}.\footnote{My
examples assume that all farmers are male.}
\ex.\label{e:donkey}
    Every farmer who owns a donkey beats it.\par

Before examining its variations, a derivation of~\eqref{e:donkey} itself is
in order.  The critical lexical items are \CCSHANphrase{every} and \CCSHANphrase{who}.
Given the denotation of \CCSHANphrase{a} specified above, we expect
\CCSHANphrase{every} to have the semantic type
\[
    (e\CCSHANOut e \CCSHANtoF \sigma\CCSHANOut 1) \CCSHANtoF (\sigma\CCSHANOut e \CCSHANtoF \sigma'\CCSHANOut 1) \CCSHANtoF 1.
\]
The same semantic type is also expected for other strongly quantificational
elements, such as \CCSHANphrase{most}.  Following standard treatment in dynamic
predicate logic, I let \CCSHANphrase{every} denote%
  \footnote{
    This meaning gives the donkey antecedent universal quantificational
    force; in other words, it makes~\eqref{e:donkey} mean that every farmer
    who owns a donkey beats every donkey he owns.  As Schubert and
    Pelletier~\cite{schubert-generically} and others point out, sometimes
    the donkey antecedent seems to take existential quantificational force
    instead.  For example, \Next[a] naturally means \Next[b].
    \ex.\a. Every man who had a dime put it in the meter.
        \b. Every man who had a dime put a dime in the meter.\par
    I leave it for future work to account for this variation within the
    present framework.  One possible solution is to posit alternative
    denotations for \CCSHANphrase{every}.  Another is to treat \CCSHANphrase{it} as a
    paycheck pronoun that repeats the existential force of \CCSHANphrase{a dime},
    effectively implementing the paraphrase in~\Last[b].

    Related is the \emph{proportion problem}, noted by
    Kadmon~\cite{kadmon-unique} and others: Each sentence in~\Next has a
    different truth condition.
    \ex.\a. Most farmers who own a donkey beat it.
        \b. Most donkeys owned by a farmer are beaten by him.
        \c. Mostly, when a farmer owns a donkey, he beats it.\par
    The type constructor $\CCSHANOut$ is not symmetric; it distinguishes between
    the individual that participates immediately in predicate-argument
    combination and any additional output available as candidate
    antecedents for future anaphora.  Thus the differences in~\Last can
    easily be modeled here by positing a natural denotation for
    \CCSHANphrase{most}.}
\[
    \CCSHANEvery = \CCSHANfun{p} \CCSHANfun{q}
        \bigl\{\, * \mid
               \CCSHANForall{s:\sigma} \CCSHANForall{v:e}
               \langle s,*\rangle \in p(\langle v,v\rangle) \CCSHANImplies
               \CCSHANExists{s':\sigma'}
               \langle s',*\rangle \in q(\langle s,v\rangle)
        \,\bigr\}.
\]
As for \CCSHANphrase{who}, since I will only consider relative clauses with
subject extraction in this paper, the following denotation is sufficient.%
  \footnote{
    It is no accident that the set comprehension notation used to specify
    this meaning is reminiscent of the list or monad comprehension notation
    used to express evaluation sequencing in programming
    languages~\cite{wadler-comprehending}.}
\begin{align*}
    \CCSHANWho&\CCSHANeqishcolon (\sigma_2\CCSHANOut e \CCSHANtoF \sigma_3\CCSHANOut 1)
                \CCSHANtoF (\sigma_1\CCSHANOut e \CCSHANtoF \sigma_2\CCSHANOut 1)
                \CCSHANtoF (\sigma_1\CCSHANOut e \CCSHANtoF \sigma_3\CCSHANOut 1)\\
        &= \CCSHANfun{p} \CCSHANfun{q} \CCSHANfun{\langle s_1,v\rangle}
           \bigl\{\, \langle s_3,*\rangle
                \mid \langle s_2,*\rangle \in q(\langle s_1,v\rangle) \CCSHANcomma
                     \langle s_3,*\rangle \in p(\langle s_2,v\rangle)
           \,\bigr\}.
\end{align*}

We are now ready to derive~\eqref{e:donkey}:
\begin{gather*}
    \CCSHANFarmer\CCSHANcomma\CCSHANDonkey: e\CCSHANtoF 1,              \qquad
    \CCSHANOwn\CCSHANcomma\CCSHANBeat     : e\CCSHANtoF e\CCSHANtoF 1,        \qquad
    \CCSHANIt                 : e\CCSHANIn e,
\\
    x =
    \CCSHANWho\bigl(\CCSHANgOut_{1,0}(\CCSHANgOut_{0,1}(\CCSHANOwn))(\CCSHANA(\CCSHANgOut_{0,0}(\CCSHANDonkey)))\bigr)
        \bigl(\CCSHANgOut_{0,0}(\CCSHANFarmer)\bigr)
        : e\CCSHANOut e \CCSHANtoF e\CCSHANOut e\CCSHANOut 1,
\\
    y =
    \CCSHANgOut_{1,0}(z(\CCSHANgOut_{1,0}(\CCSHANBeat)))(\CCSHANIt)
        : e\CCSHANOut e\CCSHANOut e \CCSHANtoF e\CCSHANOut e\CCSHANOut 1,     \qquad
    \CCSHANEvery(x)(y) : 1.
\end{gather*}

\subsection{Weak crossover}

Variable-free dynamic semantics accounts for the ``donkey weak crossover''
contrasts in \eqref{e:static-wco} and~\eqref{e:dynamic-wco} in roughly the
same way regular variable-free semantics accounts for the weak crossover
contrast in~\eqref{e:static-wco}
alone~\cite[\S2.2.3]{jacobson-variable-free}.%
\ex.\label{e:static-wco}
    \a. \label{e:static-wco-good}%
        Every farmer\CCSHANIndex{i} who owns a donkey loves his\CCSHANIndex{i} mother.
    \b.*\label{e:static-wco-bad}%
        His\CCSHANIndex{i} mother loves every farmer\CCSHANIndex{i} who owns a donkey.\par
\ex.\label{e:dynamic-wco}
    \a. \label{e:dynamic-wco-good}%
        Every farmer who owns a donkey\CCSHANIndex{i} loves the woman who beats it\CCSHANIndex{i}.
    \b.*\label{e:dynamic-wco-bad}%
        The woman who beats it\CCSHANIndex{i} loves every farmer who owns a donkey\CCSHANIndex{i}.\par
More specifically, binding is disallowed in~\eqref{e:static-wco-bad}
and~\eqref{e:dynamic-wco-bad} because $z$ forces the binder---or, in the
case of donkey anaphora, the NP containing the binder---to
c\nobreakdash-command the bindee.  For the disallowed binding
configurations to be possible, the grammar would need an alternative
binding operation%
\begin{equation}
\label{e:s}
\begin{split}
s &\CCSHANeqishcolon
     \bigl((\sigma\CCSHANOut\beta) \CCSHANtoF           \alpha  \CCSHANtoF \gamma\bigr) \CCSHANtoF
     \bigl((\sigma\CCSHANOut\beta) \CCSHANtoF (\sigma\CCSHANIn\alpha) \CCSHANtoF \gamma\bigr)\\
  &= \CCSHANfun{f} \CCSHANfun{\langle s,u\rangle} \CCSHANfun{v} f(\langle s,u\rangle)(v(s)).
\end{split}
\end{equation}

\subsection{Functional questions}

Regular variable-free semantics derives the functional question-answer pair
in~\eqref{e:static-fq} and predicts the weak crossover violation
in~\eqref{e:static-fq-wco} under natural analyses of
extraction~\cite[\S3.1\nobreakdash--2]{jacobson-variable-free}.
Variable-free dynamic semantics further derives the ``donkey functional''
question-answer pair in~\eqref{e:dynamic-fq} while predicting the donkey
weak crossover violation in~\eqref{e:dynamic-fq-wco}.  The answers are all
of type $e\CCSHANIn e$.%
\ex.\label{e:static-fqs}
    \a. Who does every farmer\CCSHANIndex{i} who owns a donkey love? His\CCSHANIndex{i} mother.
        \label{e:static-fq}
    \b. Who loves every farmer\CCSHANIndex{i} who owns a donkey? ?His\CCSHANIndex{i} mother.
        \label{e:static-fq-wco}\par
\ex.\label{e:dynamic-fqs}
    \a. Who does every farmer who owns a donkey\CCSHANIndex{i} love? The woman who~beats~it\CCSHANIndex{i}.%
        \label{e:dynamic-fq}
    \b. Who loves every farmer who owns a donkey\CCSHANIndex{i}? ?The woman who beats it\CCSHANIndex{i}.%
        \label{e:dynamic-fq-wco}\par

\subsection{Across-the-board binding}

Typical compositional analyses of right node raising allow variable-free
semantics to
predict~\eqref{e:static-atb}~\cite[\S3.4]{jacobson-variable-free};
variable-free dynamic semantics further predicts~\eqref{e:dynamic-atb}.
The conjuncts are all of type $(e\CCSHANIn e)\CCSHANtoF 1$.%
\ex.\label{e:static-atb}
    Every farmer\CCSHANIndex{i} who owns a donkey loves---but no farmer\CCSHANIndex{j}
    who owns a donkey wants to marry---his\CCSHANIndex{i/j/*k} mother.\par
\ex.\label{e:dynamic-atb}
    Every farmer who owns a donkey\CCSHANIndex{i} loves---but no farmer who owns
    a donkey\CCSHANIndex{j} wants to marry---the woman who beats it\CCSHANIndex{i/j/*k}.\par

\subsection{Sloppy readings}

Because VPs can take semantic type $e\CCSHANOut e\CCSHANOut e\CCSHANtoF 1$ in variable-free
dynamic semantics, it is straightforward to derive ``sloppy'' readings for
sentences involving VP ellipsis~\eqref{e:dynamic-vpe} or association with
focus~\eqref{e:dynamic-focus}.%
\ex.\label{e:dynamic-vpe}
    \a. Every East Coast farmer who owns a donkey beats it, but no West
        Coast farmer who owns a donkey does.
    \b. Every farmer who owns a donkey showed it to his mother, but no
        farmer who owns a horse did.\par
\ex.\label{e:dynamic-focus}
    Only the farmer who owns \textsc{this} donkey beats it.\par

\subsection{Antecedent accessibility}

Muskens~\cite{muskens-combining} notes the contrast
in~\eqref{e:accessibility}.%
\ex.\label{e:accessibility}
    \a. \label{e:accessibility-good}No girl walks.
    \b.*\label{e:accessibility-bad}No girl\CCSHANIndex{i} walks. If she\CCSHANIndex{i}
        talks, she\CCSHANIndex{i} talks.\par
The second sentence in~\eqref{e:accessibility-bad} accesses a discourse
referent \CCSHANphrase{she} but is a tautology.  Muskens's semantic theory is one
of many that cannot account for this contrast without introducing
representational constraints.

In variable-free semantics, types keep track of the discourse referents
free in each phrase.  For example, the second sentence
in~\eqref{e:accessibility-bad} has type $e\CCSHANIn1$, not $1$, even though it
does denote a constant function.  The theory here thus
blocks~\eqref{e:accessibility-bad} and accounts for Muskens's observation
while remaining non-representational.

\section{Discussion}

From a theoretical perspective, variable-free dynamic semantics is
appealing for the same reasons variable-free semantics and dynamic
semantics are appealing---because it preserves direct compositionality,
eliminates assignment functions, models updates to information states, and
treats donkey anaphora.

There have been numerous proposals to combine Montague grammar with dynamic
semantics, including ones by Groenendijk and
Stokhof~\cite{groenendijk-dynamic-montague},
Muskens~\cite{muskens-combining}, Kohlhase, Kuschert, and
M\"uller~\cite{kohlhase-dynamic}, and
van~Eijck~\cite{van-eijck-incremental}.  The variable-free approach here
clearly shares the outlook of van~Eijck's work, which uses de~Bruijn
indexing to eliminate variable indices from dynamic reasoning.  By
comparison, the theory here is more prominently guided by types, using them
to record the \emph{computational side effects}~\cite{moggi-abstract} of
nondeterminism, input, and output that are incurred during dynamic
interpretation.  This type system seems related to but more simplistic than
Fernando's proof-theoretic semantics~\cite{fernando-type}; for example,
there are no dependent types.

I have tried to draw an analogy between incoming assignments ($\CCSHANIn$) and
outgoing ones ($\CCSHANOut$), characterizing with similar combinatorics the
former with the exponential functor and the latter with the product
functor.  The success of the analogy---between, for example, the type-shift
operations $\CCSHANgIn$ and $\CCSHANgOut$---suggests that the same combinatorics may be
applicable to an even wider range of linguistic phenomena.  For instance,
note that Hendriks's Argument Raising
operation~\cite[\S1.4.2]{hendriks-studied} follows from the functoriality
of $(\CCSHANstuff\to\sigma_1)\to\sigma_2$ for any $\sigma_1$ and~$\sigma_2$, in
the same way $\CCSHANgIn$ and $\CCSHANgOut$ follow from the functoriality of
$\sigma\CCSHANIn\CCSHANstuff$ and $\sigma\CCSHANOut\CCSHANstuff$, respectively, for any~$\sigma$.

\bibliographystyle{plain}
\bibliography{ccshan.bib}

\end{document}